\documentclass[runningheads]{llncs}

\usepackage[T1]{fontenc}
\usepackage{booktabs}
\usepackage{graphicx}
\usepackage{multirow}
\usepackage{url}
\usepackage{xcolor}
\newcommand{\meansd}[2]{\shortstack{#1\\[-1pt]{\scriptsize\color{gray}(#2)}}}
\newcommand{\ethicsname}{Ethics statement.}

\begin{document}

\title{Assessing nnU-Net Generalization across Brain Tumor Populations in
BraTS-GoAT 2026}
\titlerunning{Assessing nnU-Net Generalization in BraTS-GoAT 2026}

\author{Tristan Kirscher\inst{1,2} \and
Vivian Metzger\inst{2} \and
Philippe Meyer\inst{1,2} \and
Xavier Coubez\inst{1,2}}
\authorrunning{T. Kirscher et al.}
\institute{ICube Laboratory, CNRS UMR 7357, University of Strasbourg,
Strasbourg, France \and
CLCC Institut Strauss, Strasbourg, France\\
\email{tristan.kirscher@unistra.fr}}

\maketitle

\begin{abstract}
BraTS-GoAT evaluates tumor segmentation across heterogeneous populations. We
trained a conventional 3D nnU-Net on 1,351 labeled cases using
five-fold cross-validation and 1,000 epochs per fold. The final predictor
averaged all folds and applied test-time mirroring. On pooled official
validation, global DSC values were 0.7805, 0.8288, and 0.8854 for enhancing tumor
(ET), tumor core (TC), and whole tumor (WT). Under matched fold-0 inference, mean regional
Dice decreased from 0.9058 on source out-of-fold (OOF) cases to 0.8310 on pooled
validation (difference $-0.0747$).
Mirroring gave small single-fold gains but no clear ensemble benefit; a
residual-encoder alternative reached 0.8282 mean Dice.
In labeled OOF predictions, failure cases had substantially smaller reference
ET volumes; after adjustment for ET and WT volume, lower Dice remained
associated with more disconnected ET components and a smaller fraction of ET
contained in the largest component.

\end{abstract}

\section{Introduction}

Automatic brain-tumor segmentation supports quantitative assessment, treatment
planning, and follow-up, but robustness across populations and acquisition
protocols remains challenging. BraTS established a common evaluation framework
for multisequence MRI tumor segmentation
\cite{menze2015brats,baid2021brats,bakas2017advancing}. The BraTS
Generalizability Across Tumors (BraTS-GoAT) task extends this objective across
adult glioma, glioma from sub-Saharan Africa, meningioma, brain metastases, and
pediatric brain tumors
\cite{adewole2023africa,labella2023men,moawad2023mets,kazerooni2023peds,kazerooni2024peds}.

Tumor size, multiplicity, appearance, and imaging characteristics vary across
cohorts. nnU-Net provides a self-configuring reference
\cite{isensee2021nnunet}, while previous GoAT work combined multiple
architectures with adaptive post-processing
\cite{jiang2024generalizability}. Population shifts may also interact with
nnU-Net's data-driven configuration. In pediatric CT organ segmentation,
configuration plans derived from adult dataset fingerprints have been shown to
underperform on pediatric anatomy, particularly for small structures
\cite{kirscher2025psat}.
Motivated by these observations, we evaluated a fixed
nnU-Net protocol, ensembling, mirroring, and a residual encoder, and then
analyzed out-of-fold (OOF) failures. Because validation population labels were
unavailable, all target-domain analyses were necessarily pooled.

\section{Materials and Methods}

\subsection{Task, data, and target regions}

\begin{figure}[t]
\centering
\includegraphics[width=\textwidth]{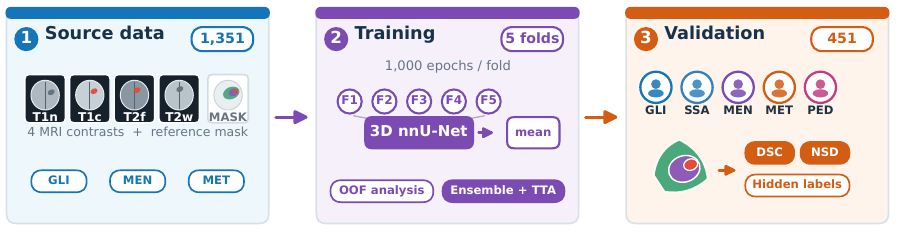}
\caption{Study design. One protocol was fitted to labeled source data and
evaluated on the pooled validation cohort without population-specific
adaptation.}
\label{fig:design}
\end{figure}

We used only challenge data, without external data or pretrained weights.
The pooled, anonymized training set contained 1,351 labeled cases drawn from three
source populations: adult glioma (GLI), meningioma (MEN), and brain metastases
(MET). Verified per-case cohort labels were unavailable. Inputs were
non-contrast T1-weighted (T1n), contrast-enhanced T1-weighted (T1c), T2-FLAIR
(T2f), and T2-weighted (T2w) MRI. The pooled validation set contained 451 cases
from the three source populations and two additional target populations:
glioma from sub-Saharan Africa (SSA) and pediatric brain tumors (PED).
Validation reference labels and the case-to-population mapping were hidden.
Figure~\ref{fig:design} summarizes the resulting source and pooled-validation
data flow.

Class 1 was NCR/NET; class 2, edema/invaded tissue; and class 3, ET. NCR/NET
is the necrotic and non-enhancing core. Evaluation used ET $\{3\}$, tumor core (TC)
$\{1,3\}$, and whole tumor (WT) $\{1,2,3\}$. Each output was a NIfTI label map
in the input image space.

\subsection{Preprocessing and network configuration}

We used nnU-Net v2.6.2 with its selected \texttt{3d\_fullres} configuration.
Modalities were resampled to 1-mm isotropic spacing and z-score normalized
within the nonzero mask. The six-stage 3D PlainConvUNet used
$128\times160\times112$ patches, batch size 2, encoder widths 32, 64, 128,
256, 320, and 320, and overlapping sigmoid outputs for WT, TC, and ET. The
encoder stages and the five decoder stages each used two
$3\times3\times3$ convolutions.
Downsampling was
$2\times2\times2$ except for a $2\times2\times1$ deepest transition.
Convolutions were followed by instance normalization and leaky ReLU.

\subsection{Training protocol}

A model was trained from scratch for each deterministic fold for 1,000 epochs (250
training and 50 validation iterations per epoch) using the standard nnU-Net
sum of region-based Dice and binary-cross-entropy losses, deep supervision, foreground
oversampling, and augmentations. Optimization used stochastic gradient descent,
Nesterov momentum 0.99, initial learning rate $10^{-2}$, weight decay
$3\times10^{-5}$, and polynomial decay. We retained the epoch-1,000 final
checkpoint for every fold.
In each cross-validation rotation, 270 or 271 source cases formed the held-out
fold used for internal validation; this internal split was distinct from the
hidden challenge validation cohort. Foreground oversampling was 0.33. Standard
augmentations covered geometry, noise and blur, intensity, simulated
resolution, gamma, and mirroring.
Checkpoints were not selected using challenge validation scores.
No hyperparameter search was performed: preprocessing, architecture, patch and
batch sizes, loss, optimizer, and schedule followed the nnU-Net plan and stock
trainer. Training plus final OOF validation took 12.1--12.5 h per fold (mean
12.3 h) on one NVIDIA A100 40-GB GPU. Peak allocated VRAM was not instrumented.

\subsection{Five-fold inference and test-time augmentation}

The final configuration averaged folds 0--4 and nnU-Net's mirrored predictions
over spatial axes $(0,1,2)$. Probability maps were averaged before conversion
of overlapping regions to atomic labels, and nnU-Net restored the input
geometry. This was a cross-validation ensemble---members were trained on
different 80\% subsets---rather than a deep ensemble trained repeatedly on the
full set. It reused the models required for OOF analysis; recent controlled
evidence found no consistent OOD winner between the two constructions and
recommends reporting them explicitly \cite{kirscher2026folds}. No
connected-component post-processing or tuned region threshold was applied.

\subsection{Alternative architecture and ablation design}

The alternative five-fold ResidualEncoderUNet-L used
$160\times192\times160$ patches, batch size 3, and the same trainer and
1,000-epoch schedule. It changed the automatically planned encoder and patch
geometry while preserving the split and inference procedure.
The five submitted configurations compared fold 0 with five-fold probability
averaging, with and without mirroring, and the residual alternative. Challenge
validation results were used only for this post hoc comparison, not for
checkpoint selection or parameter tuning.

\subsection{Validation metrics}

The server returned per-case global semantic Dice similarity coefficient
(DSC), normalized surface Dice (NSD), and 95th-percentile Hausdorff distance
(HD95).
DSC and NSD are the announced ranking metrics; the organizers specified
$\tau=1$ for final NSD scoring without specifying its unit. The tolerance and
unit used by the validation server were not disclosed; consequently, we report
its NSD values without assigning a millimeter tolerance.
We retain HD95 as a complementary diagnostic \cite{maierhein2024metricsreloaded}.
Paired contrasts used finite shared cases and 10,000-resample percentile
bootstrap intervals. We report effect estimates rather than hypothesis tests
because configurations were selected post hoc.
Table~\ref{tab:global} contains marginal means and case-level standard
deviations of the server-returned metrics over finite cases. Paired contrasts
instead exclude non-finite cases jointly for each comparison and metric.

\subsection{Pooled source--validation comparison}

For a model-matched comparison, we applied the same fold-0
checkpoint with mirroring to its 271 held-out source cases and to the official
validation cohort. We defined the performance difference as pooled-validation
DSC minus source OOF DSC.
Source and validation cases were resampled independently, and the regional mean was
recomputed in each of 10,000 bootstrap replicates. This descriptive difference
combines evaluator implementation, case mixture, and possible distributional
effects; it is not an estimate of pure population shift.

\subsection{Out-of-fold failure analysis}

Each training case was predicted by its excluding fold. The outcome was strict
mean ET/TC/WT Dice; 1,343 cases had three finite values.
Failures were defined as cases in the bottom decile of the strict mean regional
Dice distribution. Robustness to this threshold was assessed using alternative
bottom-5\% and bottom-20\% definitions.

Reference-mask features comprised volume, 26-connected components of at least
10 voxels (to suppress isolated voxel-scale fragments), largest-component
fraction, and ET/WT ratio. At 1-mm isotropic spacing, voxel counts were divided
by 1,000 to report volumes in mL. We used rank-biserial effects for group contrasts and partial
Spearman correlations, computed from rank residuals after adjustment for the
stated volume covariates. Benjamini--Hochberg correction was applied separately
to the morphology screen and four targeted follow-up tests.
Within each of 2,000 fold-stratified bootstrap resamples, ranks and residual
models were re-estimated.
Intervals quantify case-sampling uncertainty with the trained segmentation
models fixed; they omit retraining and hidden-population-mixture uncertainty.
All failure analyses remain exploratory because hypotheses and estimates used
the same cases and folds had overlapping training sets.

\subsection{Container and reproducibility controls}

The offline Linux/AMD64 container pins nnU-Net 2.6.2, PyTorch 2.6.0, CUDA
12.4, and cuDNN 9. It reads \texttt{/input}, writes one flat NIfTI map per case
to \texttt{/output}, and validates modalities, cases, and geometry against T1n.
Code, nnU-Net plans, container files, inference configuration, validation
checks, and analysis scripts are available at
\url{https://github.com/Kirscher/BraTS2026}. A versioned bundle of the five
epoch-1,000 checkpoints is linked from the repository. Challenge data,
predictions, and official score exports are not redistributed; analyses that
depend on them therefore support inspection but not regeneration from a clean
clone. Five-fold mirrored inference over the 451-case validation archive took
1 h 55 min on one NVIDIA Quadro RTX 6000 24-GB GPU (15.3 s/case); peak
allocated VRAM was not instrumented.

\section{Results}

\subsection{Internal five-fold cross-validation}

The unweighted mean of the mean foreground Dice values in the five final
nnU-Net fold summaries was 0.9022 (SD 0.0073; range 0.8913--0.9090). Pooled
per-case OOF regional Dice was
0.8674 (SD 0.1962) for ET ($n=1{,}343$), 0.9125 (0.1573) for TC
($n=1{,}350$), and 0.9267 (0.1081) for WT ($n=1{,}351$). The strict
complete-case mean was 0.9034 (SD 0.1275; $n=1{,}343$).

\subsection{Official validation results}

All five 451-case archives were accepted. The server omitted case 02845 from
every per-case score file, leaving the same 450 rows. The five-fold TTA model
had the highest mean DSC (0.8315) and NSD (0.5164), while fold-0 TTA reached
0.8310 and 0.5140. The residual configuration had lower DSC point estimates in
every region, but its paired intervals against PlainConv included zero
(Table~\ref{tab:global}).

\begin{table}[t]
\centering
\caption{Official pooled validation results: mean (case-level SD) over finite
server-returned values. All configurations share 450 scored cases from 451-case
archives; the server omitted case 02845, and finite counts vary for empty ET/TC
regions. Higher is better for DSC and NSD, and lower for HD95; bold marks the
best mean per column. ``+TTA'' denotes mirroring.}
\label{tab:global}
\small
\setlength{\tabcolsep}{2pt}
\begin{tabular}{lcccccc}
\toprule
& \multicolumn{3}{c}{Global DSC $\uparrow$}
& \multicolumn{3}{c}{Global NSD $\uparrow$}\\
\cmidrule(lr){2-4}\cmidrule(lr){5-7}
Configuration & ET & TC & WT & ET & TC & WT\\
\midrule
PlainConv, fold 0
& \meansd{0.7738}{.297} & \meansd{0.8246}{.250} & \meansd{0.8859}{.166}
& \meansd{0.5392}{.257} & \meansd{0.5025}{.274} & \meansd{0.4825}{.198}\\
PlainConv, fold 0 +TTA
& \meansd{\textbf{0.7811}}{.292} & \meansd{0.8251}{.254} & \meansd{\textbf{0.8869}}{.167}
& \meansd{0.5476}{.256} & \meansd{0.5065}{.276} & \meansd{0.4880}{.202}\\
PlainConv, folds 0--4
& \meansd{0.7777}{.299} & \meansd{0.8269}{.249} & \meansd{0.8855}{.174}
& \meansd{0.5472}{.263} & \meansd{0.5082}{.281} & \meansd{0.4907}{.204}\\
PlainConv, folds 0--4 +TTA
& \meansd{0.7805}{.297} & \meansd{\textbf{0.8288}}{.247} & \meansd{0.8854}{.175}
& \meansd{\textbf{0.5495}}{.264} & \meansd{\textbf{0.5089}}{.282} & \meansd{0.4908}{.206}\\
Residual-L, folds 0--4 +TTA
& \meansd{0.7776}{.306} & \meansd{0.8249}{.259} & \meansd{0.8821}{.190}
& \meansd{0.5476}{.275} & \meansd{0.5063}{.291} & \meansd{\textbf{0.4910}}{.222}\\
\bottomrule
\end{tabular}

\vspace{3pt}

\begin{tabular}{lccc}
\toprule
& \multicolumn{3}{c}{Global HD95 $\downarrow$}\\
\cmidrule(lr){2-4}
Configuration & ET & TC & WT\\
\midrule
PlainConv, fold 0 & \meansd{41.67}{109.60} & \meansd{20.19}{67.78} & \meansd{12.75}{50.65}\\
PlainConv, fold 0 +TTA & \meansd{\textbf{39.87}}{107.69} & \meansd{20.65}{69.68} & \meansd{13.43}{53.33}\\
PlainConv, folds 0--4 & \meansd{40.35}{108.65} & \meansd{\textbf{18.83}}{65.45} & \meansd{\textbf{12.70}}{50.62}\\
PlainConv, folds 0--4 +TTA & \meansd{41.18}{110.04} & \meansd{19.35}{67.51} & \meansd{13.58}{53.48}\\
Residual-L, folds 0--4 +TTA & \meansd{44.31}{114.57} & \meansd{23.13}{77.06} & \meansd{16.62}{65.05}\\
\bottomrule
\end{tabular}
\end{table}

\subsection{Quantitative ablation effects}

Paired complete-case effects need not equal differences between the marginal
means in Table~\ref{tab:global}. For fold 0, TTA increased
ET and WT DSC by 0.0021 (95\% interval
0.0003--0.0039) and 0.0011 (0.00003--0.0021); TC changed by 0.0005
($-0.0022$--0.0032). NSD gains were 0.0040--0.0055. After five-fold
ensembling, every DSC and NSD change was within 0.0008 of zero and all
intervals included zero.

Five folds gave no clear paired DSC improvement over fold 0. Without TTA, ensembling
increased NSD by 0.0056--0.0082; with TTA, intervals included zero. Residual
minus conventional five-fold TTA DSC changes were $-0.0047$, $-0.0039$, and
$-0.0033$, with intervals including zero.

An exploratory cross-fitted OOF sweep over ET probability thresholds and
minimum component sizes did not improve the default operating point: ET DSC
changed by $-0.00072$ (95\% interval $-0.00122$ to $-0.00032$), and ET NSD by
$-0.00225$ ($-0.00334$ to $-0.00137$). Because calibration-fold models had
seen some evaluated cases during training, this sweep was not fully nested; it
did not inform the final pipeline.

\subsection{Descriptive pooled source--validation gap}

Under matched fold-0 TTA inference, mean regional DSC decreased from 0.9058
internally to 0.8310 on pooled validation ($-0.0747$, 95\% interval
$-0.0976$ to $-0.0523$; Table~\ref{tab:gap}). TC and ET had the largest gaps.
Hidden labels prevent population-specific attribution.

\begin{table}[p]
\centering
\caption{Matched fold-0 TTA source--validation DSC ($n=271$ source; $n=450$
validation). Source and validation cases were independently resampled 10,000
times; intervals report validation minus source.}
\label{tab:gap}
\small
\setlength{\tabcolsep}{5pt}
\begin{tabular}{lccc}
\toprule
Region & Source OOF & Pooled validation & Validation $-$ source [95\% interval]\\
\midrule
ET & 0.8666 & 0.7811 & $-0.0856$ [$-0.1213,-0.0490$]\\
TC & 0.9174 & 0.8251 & $-0.0923$ [$-0.1217,-0.0631$]\\
WT & 0.9333 & 0.8869 & $-0.0464$ [$-0.0642,-0.0292$]\\
Mean & 0.9058 & 0.8310 & $-0.0747$ [$-0.0976,-0.0523$]\\
\bottomrule
\end{tabular}
\end{table}

\begin{figure}[p]
\centering
\includegraphics[width=\textwidth]{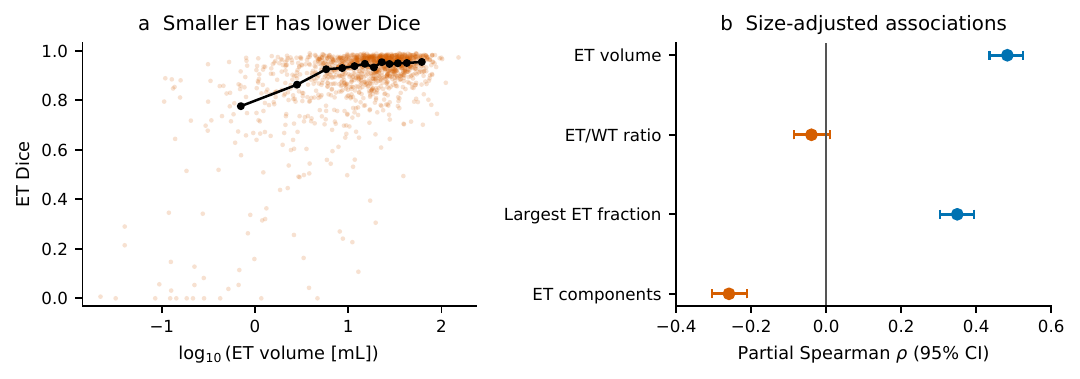}
\caption{Exploratory OOF failures. \textbf{a} ET Dice versus reference ET
volume; black points connect 12 equal-frequency bin medians. \textbf{b}
Partial Spearman associations with mean regional Dice: ET volume adjusted for
WT volume; the other features adjusted for ET and WT volume. Intervals are
fold-stratified 95\% bootstrap intervals.}
\label{fig:failure}
\end{figure}

\subsection{Failure conditions in labeled out-of-fold predictions}

The bottom-decile threshold (mean Dice 0.7877) identified 135 of 1,343 cases.
Failure cases had a median reference ET volume of 1.04 versus 18.66 mL for the non-failure cases
(rank-biserial effect $-0.808$, $q=2.34\times10^{-52}$).
The rank-biserial effect comparing reference ET volume between failure and
non-failure cases remained negative under the bottom-5\%, bottom-10\%, and
bottom-20\% failure definitions ($-0.938$, $-0.808$, and $-0.702$,
respectively).

\begin{figure}[!t]
\centering
\includegraphics[width=\textwidth]{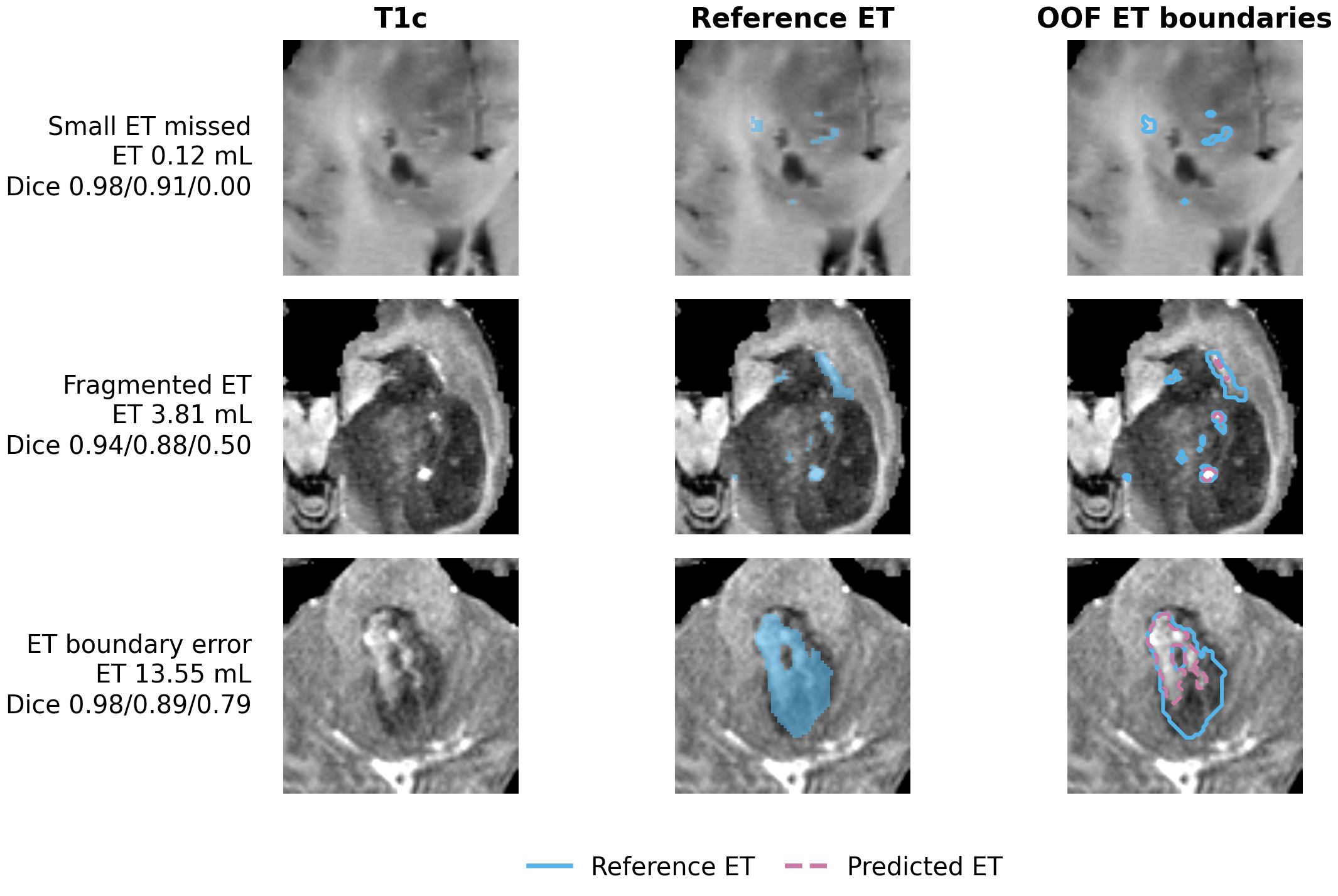}
\caption{Illustrative OOF ET failures on T1c: a small missed ET
(GoAT-01055, fold 3), incomplete fragmented ET (GoAT-00330, fold 3), and local
boundary under-segmentation (GoAT-01029, fold 1). Dice values, recomputed from
the displayed 3D masks, are for WT/TC/ET respectively. Selected examples do not reflect
failure frequencies or population effects and illustrate only these error modes.}
\label{fig:qualitative}
\end{figure}

After adjustment for WT volume, ET volume remained associated with mean Dice
($\rho=0.483$, 95\% bootstrap interval 0.436--0.526). Conditional on ET and WT
volume, a larger dominant ET component was favorable ($\rho=0.350$,
0.304--0.395), whereas more ET components were unfavorable ($\rho=-0.259$,
$-0.304$ to $-0.212$). The ET/WT volume ratio added no clear association after
size adjustment ($\rho=-0.039$, $-0.086$ to 0.010; $q=0.151$).
The associations for ET volume, largest ET component fraction, and ET component
count each had $q<10^{-20}$, although all four targeted analyses remain
exploratory. Figure~\ref{fig:failure} summarizes the volume effect and the
size-adjusted associations.
Three selected cases illustrate missed small ET, incomplete fragmented ET,
and local boundary errors despite accurate WT (Fig.~\ref{fig:qualitative}).

\section{Discussion and Conclusion}

Validation DSC was lower than source OOF DSC. Mirroring helped fold 0 but not
the ensemble; neither the residual configuration nor the ET sweep clearly
improved performance. Five-fold TTA was selected post hoc by mean validation
DSC and NSD and does not establish general superiority.

OOF failures concentrated in small and fragmented ET, but source-only mask
associations are neither causal nor deployable and may not characterize SSA or
PED. The source--validation gap combines evaluator, case-mixture, and possible
distributional effects rather than pure shift. Limitations include hidden
validation labels and population mapping, post hoc selection, an undisclosed
validation NSD tolerance, and intervals omitting retraining uncertainty.
Overall, conventional nnU-Net was a transparent baseline; the tested additions
gave no clear improvement.

\begin{credits}
\subsubsection{\ackname}
Data used in this publication were obtained as part of the Challenge project
through Synapse ID (\texttt{syn74274097}). The authors thank the BraTS 2026
organizers and contributors. This work of the Interdisciplinary Thematic
Institute HealthTech, as part of the ITI 2021--2028 program of the University
of Strasbourg, CNRS, and Inserm, was partially supported by IdEx Unistra
(ANR-10-IDEX-0002) and SFRI (STRAT'US project, ANR-20-SFRI-0012) under the
framework of the French Investments for the Future Program. The authors
acknowledge the High Performance Computing Center of the University of
Strasbourg for scientific support and access to computing resources. Part of
the computing resources was funded by the Equipex Equip@Meso project
(Programme Investissements d'Avenir) and the CPER Alsacalcul/Big Data.

\subsubsection{\ethicsname}
This study is a secondary analysis of anonymized, organizer-provided challenge
data; the authors had no access to direct patient identifiers.

\subsubsection{\discintname}
The authors have no competing interests to declare that are relevant to the
content of this article.
\end{credits}

\bibliographystyle{splncs04}
\bibliography{references}

\end{document}